\documentclass{article}
\usepackage{ijcai22}

\usepackage{times}
\usepackage{xcolor}
\usepackage{lipsum}
\usepackage{soul}
\usepackage{url}
\usepackage[hidelinks]{hyperref}
\usepackage[utf8]{inputenc}
\usepackage{caption}
\usepackage{makecell}
\usepackage{enumitem}
\usepackage{graphicx}
\usepackage{amssymb}
\usepackage{amsmath}
\usepackage{cleveref}
\usepackage{amsthm}
\usepackage{booktabs}
\usepackage{algorithm}
\usepackage{algorithmic}
\usepackage{multirow}
\newtheorem{theorem}{Theorem}
\newtheorem{lemma}{Lemma}

\crefname{equation}{}{}
\Crefname{equation}{}{}
\DeclareMathOperator*{\leqt}{\leq}

\newcommand\sbullet[1][.5]{\mathbin{\vcenter{\hbox{\scalebox{#1}{$\bullet$}}}}}

\newcommand{\ec}{K}

\newcommand{\ic}{k}

\newcommand{\fedet}{Fed-ET}

\newcommand{\lff}{f}
\newcommand{\lf}{F_\ic}
\newcommand{\gl}{\mathcal{L}}

\newcommand{\sigl}{s}

\newcommand{\sw}{\alpha}
\newcommand{\dd}{\mathcal{D}}
\newcommand{\ddh}{\widehat{\mathcal{D}}}

\newcommand{\wb}{\mathbf{w}}
\newcommand{\xb}{\mathbf{x}}

\newcommand{\bdat}{\mathcal{B}_\ic}

\newcommand{\ldat}{m_\ic}

\renewcommand*{\thefootnote}{\fnsymbol{footnote}}

\newcommand{\lr}{\eta_t}

\newcommand{\wbo}{\overline{\wb}}

\newcommand{\expt}{\mathbb{E}}

\newcommand{\defeq}{\mathrel{\vcenter{\baselineskip0.5ex \lineskiplimit0pt
                     \hbox{\scriptsize.}\hbox{\scriptsize.}}}%
                     =}

\newcommand\blfootnote[1]{%
  \begingroup
  \renewcommand\thefootnote{}\footnote{#1}%
  \addtocounter{footnote}{-1}%
  \endgroup
}

\DeclareMathOperator*{\argmin}{arg\,min} 
\DeclareMathOperator*{\argmax}{arg\,max} 

\title{Heterogeneous Ensemble Knowledge Transfer for Training Large Models in Federated Learning}

\author{Content List: Federated Learning, Ensemble Methods
}

\author{
Yae Jee Cho$^{1,2}$\footnote{Work done while at Microsoft Research. Corresponding author email: \{yaejeec@andrew.cmu.edu\}.}
\and
Andre Manoel$^1$\and
Gauri Joshi$^2$\and
Robert Sim$^1$ \And
Dimitrios Dimitriadis$^1$
\affiliations
$^1$Microsoft Research, $^2$Carnegie Mellon University}

\begin{document}

\maketitle

\begin{abstract}
  Federated learning (FL) enables edge-devices to collaboratively learn a model without disclosing their private data to a central aggregating server. Most existing FL algorithms require models of identical architecture to be deployed across the clients and server, making it infeasible to train large models due to clients' limited system resources. In this work, we propose a novel ensemble knowledge transfer method named \fedet~in which small models (different in architecture) are trained on clients, and used to train a larger model at the server. Unlike in conventional ensemble learning, in FL the ensemble can be trained on clients' highly heterogeneous data. Cognizant of this property, \fedet~uses a weighted consensus distillation scheme with diversity regularization that efficiently extracts reliable consensus from the ensemble while improving generalization by exploiting the diversity within the ensemble. We show the generalization bound for the ensemble of weighted models trained on heterogeneous datasets that supports the intuition of \fedet. Our experiments on image and language tasks show that \fedet~significantly outperforms other state-of-the-art FL algorithms with fewer communicated parameters, and is also robust against high data-heterogeneity. \blfootnote{Accepted to the proceedings of the 31st International Joint Conference on Artificial Intelligence (IJCAI 2022)}
\end{abstract}
\renewcommand*{\thefootnote}{\arabic{footnote}}
\section{Introduction}
Moving both data collection and model training to the edge, federated learning (FL) has gained much spotlight since it was introduced~\cite{mcmahan2017communication}. In FL, a number of edge-devices (clients), like cell-phones or IoT devices, collaboratively train machine learning models without explicitly disclosing their local data. Instead of communicating their data, the clients locally train their models, and send model updates periodically to the aggregating  server. The two distinctive challenges in FL are that clients can have i)~limited system resources, and ii)~heterogeneous local datasets~\cite{kairouz2019advances,bonawitz2019towards}. Many recent work in FL~\cite{wang2021field} overlook the clients' resource constraints, using large homogeneous models on the clients and server. In practice, the clients do not have enough bandwidth or computing power to train large state-of-the-art (SOTA) models, and therefore, are restricted to train smaller and computationally lighter models. Moreover, a naive aggregation of the clients' models can hinder the convergence of the model due to high data-heterogeneity across the clients~\cite{sahu2019federated,yjc2021perfl}. Based on these constraints, the global model trained on clients can fail to work well in practice.

\begin{table*}
\centering
\begin{tabular}{lccccc}
\toprule
\multirow{2}{*}{Method} & Client Model  & Public  & Client Access to  & Server Model & Possible   \\
& Heterogeneity & Data & Public Data & Size &  Tasks \\
\midrule
FedAvg~\cite{mcmahan2017communication}   & No  & N/A & N/A & $=~$Client Model & Any    \\
FedProx~\cite{sahu2019federated}  & No  & N/A & N/A & $=~$Client Model & Any  \\
SCAFFOLD~\cite{karimireddy2019scaffold} & No  & N/A & N/A & $=~$Client Model & Any        \\
MOON~\cite{li2021moon}   & No  & Unlabeled & Required & $=~$Client Model  & Only Image     \\
FedDF~\cite{lin2020ensemble}   & Yes & Unlabeled & Not Required & $=~$Client Model & Any      \\

DS-FL~\cite{ita2021dsfl}   & Yes  & Unlabeled & Required & $=~$Client Model & Any     \\
FedGKT~\cite{chao2020GKT}  & Yes & N/A & N/A & $>~$Client Model & Only Image      \\
FedGEMS~\cite{cheng2021gems}  & Yes & Labeled & Required & $>~$Client Model & Any      \\
\textbf{Fed-ET (ours)}  & \textbf{Yes} & \textbf{Unlabeled} & \textbf{Not Required} & \textbf{$\boldsymbol{>}$ Client Model} & \textbf{Any}\\
\bottomrule
\end{tabular}
\vspace*{-0.5em}
\caption{Comparison of Related Work with \fedet}
\vspace*{-0.5em}
\label{table:comp}
\end{table*} 

A more realistic approach to learn from the resource-constrained clients in FL is by allowing different models to be deployed across clients depending on their system resources, all while training a larger model for the server. This presents a new challenge where clients return models not only trained on heterogeneous data, but also with different architecture (amongst themselves and the server). Hence, we raise the question: \textit{How can we utilize an ensemble of different models trained on heterogeneous datasets to train a larger model at the server?} We draw insight from ensemble knowledge transfer~\cite{hinton2015distillknow,li2021ensemble} to investigate this problem in our work in the FL context.

Previous studies on ensemble knowledge transfer~\cite{lan2018distillfly,hong2021ensemblerl,tran2020hydra,park2020ensemble}, propose methods to transfer knowledge from \emph{a bag of experts} to a target model, where the ensemble models are trained on similar datasets. These datasets are commonly generated from methods data augmentation or simple data shuffling. In FL, however, the models are trained on heterogeneous data distributions, where some models may show higher inference confidence than others, depending on the data sample used for knowledge transfer. Knowing which model is an \textit{expert} than the others for each data sample is imperative for effective ensemble transfer in FL -- especially when there are no hard labels for the data samples.

In this work, we propose a novel ensemble knowledge transfer algorithm for FL named \fedet~which trains a large model at the server via training small and heterogeneous models at the resource-constrained and data heterogeneous clients. Inspired by the successful usage of knowledge transfer via unlabeled public data~\cite{hinton2015distillknow,li2021ensemble}, \fedet~leverages unlabeled data to perform a bi-directional ensemble knowledge transfer between the server and client models. Unlike previous work in FL with knowledge distillation or with sole focus on tackling data-heterogeneity (see \Cref{table:comp}), \fedet~allows client model-heterogeneity while training a larger model at the server, and can be used for any classification tasks. Moreover, \fedet~does not impose any overhead to the clients nor assumes that the clients have access to additional data other than its private data. In \fedet~clients simply perform local training as in standard FL while all the other computations are done by the server. Our main contributions are:
\begin{itemize}[leftmargin=*]
    \item We propose \fedet, the first ensemble transfer algorithm for FL (to the best of our knowledge) using unlabeled data that enables training a large server model with smaller models at the clients, for any classification task.
    \item We consider the data-heterogeneity in FL by proposing a weighted consensus distillation approach with diversity regularization in \fedet~that effectively filters out \textit{experts}, showing the corresponding generalization bounds.
    \item We show \fedet's efficacy with image and language classification tasks where \fedet~achieves higher test accuracy, with more robustness against data-heterogeneity and fewer communication rounds, than other FL algorithms.
\end{itemize}

\section{Background and Related Work}
\paragraph{Ensemble Knowledge Transfer.} Knowledge transfer from an ensemble of trained models to a target model has been studied in various areas of machine learning. In \cite{lan2018distillfly}, ensemble knowledge distillation for online learning is proposed where the teacher ensembles are trained on-the-fly to simultaneously train the teachers along with the target model. In \cite{hong2021ensemblerl}, ensemble reinforcement learning is investigated where an ensemble of policies share knowledge through distillation. In \cite{tran2020hydra}, ensembles trained on shuffled data are used to transfer knowledge to a target model, and ways to utilize the diversity across these models to improve knowledge transfer are investigated. 

The previous work mentioned above, however, is not directly applicable to FL because i) the local models are trained on heterogeneous data, and ii) FL is an iterative process with only a fraction of clients participating in every communication round. Since the server sends its knowledge back to a new set of clients every round in FL, an ensemble knowledge transfer scheme should have a well defined feedback loop from the target model to the ensemble of models. As such, our proposed \fedet~induces a data-aware weighted consensus from the ensemble of models, with a feedback loop to transfer the server model's knowledge to the client models. 

\begin{figure*}[!t]
    \centering
    \includegraphics[width=1\textwidth]{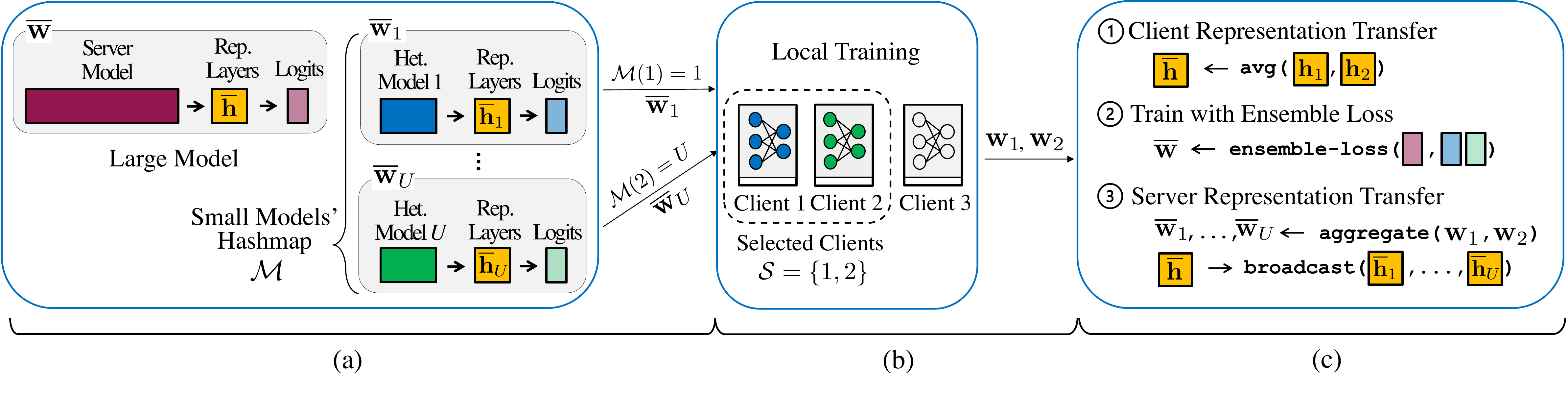} \vspace*{-2em}
    \caption{Overview of the \fedet~framework with 3 clients and $U$ small models; (a): server sends the predesignated small models in $\mathcal{M}$ to the selected clients; (b): clients perform local training and send the updates to the server; (c): server updates its large model with the received updates with FedET's primary 3 steps for ensemble transfer (see \Cref{sec:et}).}
    \label{fig:framework}
\end{figure*}

\paragraph{FL with Knowledge Distillation.} 
Several studies investigated combining FL with knowledge distillation to allow different models across clients, or to improve the server model. In \cite{ita2021dsfl}, an entropy-reduction aggregation method of the clients' logits is proposed, lowering the variance of the clients' outputs. In \cite{chao2020GKT}, FedGKT is proposed specifically for image classification, using knowledge distillation across CNNs with small CNNs at the clients and a larger CNN at the server. In \cite{li2021moon}, MOON is also proposed for image tasks, where contrastive loss is used across models of identical architecture from the server and clients to improve the server model. In \cite{lin2020ensemble}, FedDF is proposed to aggregate heterogeneous models through knowledge distillation with unlabeled public data, but it does not take data-heterogeneity into account and the server model is restricted to the clients' models. A concurrent work has proposed FedGEMS~\cite{cheng2021gems} which relies on a labeled public dataset to transfer knowledge between clients and server to train large models at the server.

The aforementioned work in FL with knowledge distillation are limited to specific scenarios such as when we have labels in the public dataset, target only image tasks, or have low data-heterogeneity across the clients. Our proposed \fedet~is not limited to these scenarios while still being able to outperform the baselines in \Cref{table:comp} as shown in our experiments. We use a weighted consensus-based distillation scheme, where clients with higher inference confidence contribute more to the consensus compared to less confident clients. We also take use of a diversity regularization term, where clients that do not follow the consensus can still transfer useful representations to the server model.


\section{Federated Ensemble Transfer: \fedet}
We propose \fedet, an ensemble knowledge transfer framework that trains a large server model with small and heterogeneous models trained on clients, using an unlabeled public dataset\footnote{Applicable datasets are accessible by the server through data generators (e.g., GAN), open-sourced repositories, or data markets.}. Concisely, \fedet~consists of the three consecutive steps: i) clients' local training and representation transfer, ii) weighted consensus distillation with diversity regularization, and iii) server representation transfer (see~\Cref{fig:framework}).

\subsection{Preliminaries}
We consider a cross-device FL setup with a $N$-class classification task where $\ec$ clients are connected to a server. Each client $\ic\in[\ec]$ has its local training dataset $\bdat$ and each data sample $\xi$ is a pair $(\xb,y)$ with input $\xb\in\mathbb{R}^d$ and label $y\in[1,N]$. Each client has its local objective $\lf(\wb)=\frac{1}{|\bdat|}\sum_{\xi \in \bdat} f(\wb,\xi)$ with $f(\wb, \xi)$ being the composite loss function. Having a large $\wb$ with identical architecture across all resource-constrained clients, as done in the standard FL framework, can be infeasible. Moreover, the local minimums $\wb_{\ic}^*,~\ic\in[1,\ec]$ minimizing $\lf(\wb)$ can be different from each other due to data-heterogeneity. \fedet~tackles these obstacles by training a large server model with data-aware ensemble transfer from the smaller models trained on clients.

Formally, we consider $U$ small and heterogeneous models at the server with $\mathcal{M}=\{1:\wbo_{1},...,U:\wbo_{U}\}$ where $\mathcal{M}$ is the hashmap with the keys $1,...,U$ as model ids, the values $\mathcal{M}[i]=\wbo_{i}\in\mathbb{R}^{n_{i}}$ as the models, and $n_{i}$ as the number of parameters for $i\in[U]$. All of the small models in $\mathcal{M}$ have a representation layer $\overline{\mathbf{h}}_i\in\mathbb{R}^{u},~i\in[U]$, which includes the classification layer, connected to the end of their different model architectures $u \ll n_{i},i\in[U]$. Each client is designated its model to use from $\mathcal{M}$ depending on its resource capability. With slight abuse of notation, we denote the model id chosen by client $k$ as $\mathcal{M}(\ic)\in[1,U]$, and the local model for that client $\ic\in[\ec]$ as $\wb_{\ic}=\wbo_{\mathcal{M}(\ic)}=\mathcal{M}[\mathcal{M}(\ic)]$ which has its respective representation layer defined as $\mathbf{h}_\ic$.

The server has its large model defined as $\wbo\in\mathbb{R}^{n}$ also with its representation layer defined as $\overline{\mathbf{h}}\in\mathbb{R}^{u}$. The large server model is assumed to be much larger than the small server models in $\mathcal{M}$, i.e., $n \gg n_{i},i\in[U]$. As shown in the following sections, the representation layers $\overline{\mathbf{h}}$ and $\mathbf{h}_\ic,\ic\in[\ec]$ are shared bidirectionally between clients and server to transfer the representations learned from their respective training. Only the server has access to an unlabeled public dataset denoted as $\mathcal{P}$. The local models $\wb_\ic,\ic\in[K]$, and large server model $\wbo$ output soft-decisions (logits) over the pre-defined number of classes $N$, which is a probability vector over the $N$ classes. We refer to the soft-decision of model $\wb_\ic$ over any input data $\xb$ in either the private or public dataset as $\sigl(\wb_\ic, \xb):\mathbb{R}^{n_{\mathcal{M}(k)}}\times(\mathcal{B}_\ic\cup\mathcal{P})\rightarrow\Delta_N$, where $\Delta_N$ stands for the probability simplex over $N$.

\subsection{Ensemble Transfer with Federated Learning} \label{sec:et}
\subsubsection{Step 1: Client Local Training \& Representation Transfer} 
For each communication round $t$, the server gets the set of $m<K$ clients, denoted as $\mathcal{S}^{(t,0)}$, by selecting them in proportion to their dataset size. The upper-subscript $(t,r)$ denotes for $t$-th communication round and $r$-th local iteration. Note that $\mathcal{S}^{(t,0)}$ is independent of the local iteration index. For each client $\ic\in\mathcal{S}^{(t,0)}$, the most recent version of its designated model $\wb_\ic^{(t,0)}=\wbo_{\mathcal{M}(\ic)}^{(t,0)}=\mathcal{M}[\mathcal{M}(k)]$ is sent from the server to the client. The clients perform local mini-batch stochastic-gradient descent (SGD) steps on their local model $\wb_\ic^{(t,0)}$ with their private dataset $\bdat,\ic\in[\ec]$. Accordingly, the clients $\ic\in\mathcal{S}^{(t,0)}$ perform $\tau$ local updates so that for every communication round their local models are updated as:
\begin{align}
\label{eqn:local_model_update}
\wb_\ic^{(t,\tau)}= \wb_\ic^{(t,0)}-\frac{\lr}{b}\sum_{r=0}^{\tau-1}\sum_{\xi \in \xi_\ic^{(t,r)}}\nabla \lff(\wb_\ic^{(t,r)}, \xi)
\end{align}
where $\lr$ is the learning rate and $\frac{1}{b}\sum_{\xi \in \xi_\ic^{(t,r)}}\nabla \lff(\wb_\ic^{(t,r)}, \xi)$ is the stochastic gradient over mini-batch $\xi_\ic^{(t,r)}$ of size $b$ randomly sampled from $\mathcal{B}_k$. After the clients $\ic\in\mathcal{S}^{(t,0)}$ finish their local updates, the models $\wb_\ic^{(t,\tau)},\ic\in\mathcal{S}^{(t,0)}$ are sent to the server. Each client has different representation layers $\mathbf{h}_\ic^{(t,\tau)}$ in their respective models $\wb_\ic^{(t,\tau)},\ic\in\mathcal{S}^{(t,0)}$. The server receives these models from the clients and updates its large model's representation layer with the ensemble models as $\overline{\mathbf{h}}^{(t,0)}=\frac{1}{m}\sum_{\ic\in\mathcal{S}^{(t,0)}}\mathbf{h}_\ic^{(t,\tau)}$. This pre-conditions the large server model with the clients' representations for Step 2 where we train the large server model with the ensemble loss.

\subsubsection{Step 2: Ensemble Loss by Weighted Consensus with Diversity Regularization}
Next, the large server model is trained via a weighted consensus based knowledge distillation scheme from the small models received from the clients. A key characteristic of the ensemble is that each model may be trained on data samples from different data distributions. Hence, some clients can be more confident than others on each of the public data samples. However, all clients may still have useful representations to transfer to the server, even when they are not very confident about that particular data sample. Hence \fedet~proposes a weighted consensus distillation scheme with diversity regularization, where the large server model is trained on the consensus knowledge from the ensemble of models while regularized by the clients that do not follow the consensus.

\paragraph{$\sbullet$~\text{Weighted Consensus:}} First we derive a reliable consensus over the ensemble of models by evaluating the variance within the logit vectors $s(\wb_\ic^{(t,\tau)},\xb),~\xb\in\mathcal{P}$ for each client $\ic\in\mathcal{S}^{(t,0)}$. We denote this variance as $\sigma^2_s(\wb_\ic^{(t,\tau)},\xb)\defeq\text{Var}(s(\wb_\ic^{(t,\tau)},\xb))$, which is the variance taken over the $N$ total probability values for the $N$-multi-class classification task. Higher $\sigma^2_s(\wb_\ic^{(t,\tau)},\xb)$ indicates a more confident client $\ic$ about how well it models data sample $\xb$, and vice-versa~\cite{Camacho-Gomez2021}. Hence, we weigh the logits from the clients with high $\sigma^2_s(\wb_\ic^{(t,\tau)},\xb)$ more heavily compared to low-variance logit clients. Formally, we set a confidence based weighted average over the logits for each data sample $\xb\in\mathcal{P}$ denoted as:
\begin{align}
\sigl^{(t,\tau)}(\xb)=\sum_{\ic\in\mathcal{S}^{(t,0)}}\alpha^{(t,\tau)}_\ic(\xb)s(\wb_\ic^{(t,\tau)},\xb)
\end{align}
where the weights are defined as:
\begin{align}
\alpha^{(t,\tau)}_\ic(\xb)=\sigma^2_s(\wb_\ic^{(t,\tau)},\xb)/\sum_{l\in\mathcal{S}^{(t,0)}}\sigma^2_s(\wb_{l}^{(t,\tau)},\xb)
\end{align}
The resulting weighted consensus logit $\sigl^{(t,\tau)}(\xb)$ efficiently derives the consensus out of the ensemble of models trained on heterogeneous datasets due to filtering out the following two main adversaries: \textbf{i) the non-experts with low intra-variance within each logit}, and \textbf{ii) overly-confident but erroneous outliers} by utilizing the power of ensemble where multiple experts contribute to the consensus.

For each data sample $\xb$ we get the most probable label from $\sigl^{(t,\tau)}(\xb)$ as:
\begin{align}
    y_s^{(t,\tau)}(\xb)=\argmax_{\text{label}\in[0,N-1]} \sigl^{(t,\tau)}(\xb)
\end{align}
The pair $(\xb,~y_s^{(t,\tau)}(\xb)),~\xb\in\mathcal{P}$ is the consensus-derived data sample from the unlabeled public dataset $\mathcal{P}$, which is then used to train the server model with the cross-entropy loss $l((\xb,~y_s^{(t,\tau)}(\xb)),\wbo^{(t,0)})$. The cross-entropy loss term used in the final ensemble loss for training the server model is:
\begin{align}
     \frac{1}{|\mathcal{P}|}\sum_{\xb\in\mathcal{P}}l((\xb, y_s^{(t,\tau)}(\xb)),\wbo^{(t,0)})
     \label{eq:crossentropy}
\end{align}

\paragraph{$\sbullet$~\text{Diversity Regularization}:} While the confidence based weighted consensus can derive a more reliable consensus from the ensemble, the diversity across the participating models is less represented. Meaningful representation information of what clients learned from their private data should be included, even when certain clients have low-confidence and may have different logits from the consensus. Encouraging diversity across models can improve the generalization performance of ensemble learning~\cite{tran2020hydra,park2020ensemble}. Hence, we gather the logits from the clients that do not coincide with the consensus, formally,
\begin{align}
\mathcal{S}_{div}^{(t,0)}(\xb)=\{l:y_s^{(t,\tau)}(\xb)\neq\argmax_{\text{label}\in[0,N-1]} &s(\wb_{l}^{(t,\tau)},\xb)\nonumber\\
&\cap~l\in\mathcal{S}^{(t,0)}
\}
\end{align}
and formulate a regularization term:
 \begin{align}
\sigl_{div}^{(t,\tau)}(\xb)=\sum_{\ic\in\mathcal{S}_{div}^{(t,0)}}\alpha(\wb_\ic^{(t,\tau)}, \xb)s(\wb_\ic^{(t,\tau)},\xb)
\end{align}
where the weights are
\begin{align}
\alpha_\ic(\xb)=\sigma^2_s(\wb_\ic^{(t,\tau)},\xb)/\sum_{l\in\mathcal{S}^{(t,0)}}\sigma^2_s(\wb_{l}^{(t,\tau)},\xb)
\end{align}
Accordingly, the diversity regularization term for the final ensemble loss is where ${KL}(\cdot,\cdot)$ is the KL-divergence loss between two logits:
\begin{align}
    KL(\sigl_{div}^{(t,\tau)}(\xb), s(\wbo^{(t,0)},\xb)) \label{eq:diversity}
\end{align}

\paragraph{$\sbullet$~\text{Final Ensemble Loss:}} Finally, combining the weighted consensus based cross-entropy loss in \cref{eq:crossentropy} with the diversity regularization in \cref{eq:diversity}, the server model is updated, in every communication round $t$, by minimizing the following objective function:
\begin{align}
    F(\wbo^{(t,0)}) = \frac{1}{|\mathcal{P}|}&\sum_{\xb\in\mathcal{P}}l((\xb, y_s^{(t,\tau)}(\xb)),\wbo^{(t,0)})\nonumber \\
    &+\lambda KL(\sigl_{div}^{(t,\tau)}(\xb), s(\wbo^{(t,0)},\xb))\label{eq-1-2}
\end{align}
To minimize the ensemble loss in \cref{eq-1-2}, instead of going through the entire dataset $\mathcal{P}$, the server model takes $\tau_s$ mini-batch SGD\footnote{Herein, SGD is depicted without loss of generality for other optimization algorithms.} steps by sampling a mini-batch $\xi_{\mathcal{P}}^{(t,r')},~r'\in[0,\tau_s-1]$ of $b_s$ data samples from $\mathcal{P}$ uniformly at random without replacement. Then, for every communication round $t$ the server performs:
\begin{align}
    \begin{aligned}
    \wbo^{(t,\tau_s)}=\wbo^{(t,0)}&-\frac{\lr}{b_s}\sum_{r=0}^{\tau_s-1}\sum_{\xi \in \xi_{\mathcal{P}}^{(t,r)}}\left[\nabla l((\xi, y_s^{(t,\tau)}(\xi)),\wbo^{(t,r)})\right.\\
    &\left.+\lambda\nabla KL(\sigl_{div}^{(t,\tau)}(\xi), s(\wbo^{(t,r)},\xi))\right]
    \end{aligned} \label{eq-1-3}
\end{align}
Note that neither the weighted ensemble term nor the diversity regularization term dominates the ensemble loss in \cref{eq-1-2} with a reasonable $\lambda$ (see \Cref{tab:divcomp}) and further because each term comes from a different set of clients. The former term is from the majority of the clients following the consensus, while the latter term is from the other clients that do not coincide with that consensus. Due to data-heterogeneity, these two different sets of clients likely change every round making it difficult for a single term to dominate the ensemble loss during training.
\begin{algorithm}[!t]
\caption{Federated Ensemble Transfer: \fedet}
\label{algo1}
\renewcommand{\algorithmicloop}{\textbf{Server do:}}
\begin{algorithmic}[1]
\STATE {\bfseries Initialize:} Hashmap of Heterogeneous Models: $\mathcal{M}=\{1:\wbo_{1}^{(0,0)},...,U:\wbo_{U}^{(0,0)}\}$; Designated Model Ids for each client $k\in[K]$: $\mathcal{M}(k)\in[1,U]$; Selected set of $m<K$ clients: $\mathcal{S}^{(0,0)}$
\STATE {\bfseries Output:} $\wbo^{(T,0)}$
\STATE {\bfseries For ${t=0,...,T-1}$ communication rounds do}:
\STATE \hspace*{0.5em} {\bfseries Clients $\ic\in\mathcal{S}^{(t,0)}$ in parallel do:}
\STATE \hspace*{1em} Receive $\wb_\ic^{(t,0)}=\wbo_{\mathcal{M}(k)}^{(t,0)}=\mathcal{M}[\mathcal{M}(k)]$ from server
\STATE \hspace*{1em} Update $\wb_\ic^{(t,\tau)}$ with \cref{eqn:local_model_update} and send it to server
\STATE \hspace*{0.5em} {\bfseries Server do:}
\STATE \hspace*{1em} Receive all updated local model $\wb_\ic^{(t,\tau)},\ic\in\mathcal{S}^{(t,0)}$.
\STATE \hspace*{1em} Transfer client representation $\overline{\mathbf{h}}^{(t,0)}\!=\!\frac{1}{m}\!\sum_{\mathcal{S}^{(t,0)}}\!\mathbf{h}_\ic^{(t,\tau)}$ 
\STATE \hspace*{1em} Update $\wbo^{(t,\tau_s)}$ with \cref{eq-1-3} and update $\mathcal{M}$ with \cref{eq-1-4} 
\STATE \hspace*{1em} Transfer server representation $\overline{\mathbf{h}}^{(t,\tau_s)}$ models in $\mathcal{M}$
\STATE \hspace*{1em} Get $\mathcal{S}^{(t+1,0)}$ by sampling in proportion to dataset sizes
\end{algorithmic}
\end{algorithm} 
\subsubsection{Step 3: Server's Representation Transfer}
Finally, we update the server's small models in $\mathcal{M}$ by aggregating the received clients' models with identical architecture by simple averaging. 
Concretely, with $\mathcal{S}_{i}^{(t,0)}\defeq\{k: k\in\mathcal{S}^{(t,0)}\cap\mathcal{M}(k)=i\},~i\in[U]$, we update $\mathcal{M}$ as 
\begin{align}
    \mathcal{M}[i]=\wbo_{i}^{(t+1,0)}=\frac{1}{\left|\mathcal{S}_{i}^{(t,0)}\right|}\sum_{k\in\mathcal{S}_{i}^{(t,0)}}\wb_\ic^{(t,\tau)},~i\in[U] \label{eq-1-4}
\end{align}
After this update, the updated $\overline{\mathbf{h}}^{(t,\tau_s)}$ from the server model $\wbo^{(t,\tau_s)}$ is transferred to all the models in $\mathcal{M}$.

\paragraph{The algorithm of \fedet.} In the preceding paragraphs, we have shown three essential components of \fedet~for federated ensemble transfer with heterogeneous models trained on heterogeneous data distributions. The complete algorithm of \fedet~can be obtained by using these components in tandem as described in \Cref{algo1}. Note that FedET is easily extendable to allow clients to define their own model architectures depending on their computing resources.

\subsection{Generalization Bound for Ensemble Transfer}
In \fedet, an ensemble of small models trained on heterogeneous data distributions is used to train a large server model for its target test data distribution. We show the generalization properties of a weighted ensemble of models trained on heterogeneous datasets in respect to the server's target distribution, supporting the weighted consensus distillation process of \fedet. We consider hypotheses $h:\mathcal{X}\rightarrow\mathcal{Y}$, with input space $\mathbf{x}\in\mathcal{X}$, label space $y\in\mathcal{Y}$, and hypotheses space $\mathcal{H}$. The loss function $l(h(\xb),y)$ measures the classification performance of $h$ for a single data point $(\xb,y)$. We define the expected loss over all data points for an arbitrary data distribution $\mathcal{D}'$ as $\mathcal{L}_\mathcal{D'}(h)=\expt_{(\xb,y)\sim\mathcal{D'}}[l(h(\xb),y)]$ for $h\in\mathcal{H}$ and assume that $\gl(h)$ is convex with range $[0,1]$. We now present the generalization bound for the server's target distribution with respect to an ensemble of weighted models trained on heterogeneous datasets below in \Cref{the1}.

\begin{theorem} With $\ec$ clients and a server for FL, we have $\mathcal{D}$ as the server's target test data distribution, and $\dd_\ic,~\ddh_\ic$ as the true and empirical data distribution, respectively, for client $\ic\in[\ec]$. We define $h_\ic=\argmin_{h}\mathcal{L}_{\dd_\ic}(h)$ and $\widehat{h}_\ic=\argmin\mathcal{L}_{\ddh_\ic}(h)$. Then, we have for the weighted ensemble of models $\sum_{i=1}^K\sw_{i}h_{\ddh_i}$ for $K$ clients with arbitrary weights $\sw_{i},i\in[K],~\sum_{i=1}^K\sw_{i}=1$, with probability at least $1-\delta$ over the choice of samples, the bound: 
\begin{align}
&\begin{aligned}
\gl_{\dd}\left(\sum_{i=1}^K\sw_{i}h_{\ddh_i}\right)\leq\sum_{i=1}^\ec\sw_{i}\gl_{\ddh_i}(h_{\ddh_i})+\sqrt{\log{\delta^{-1}}}\sum_{i=1}^\ec\frac{\sw_{i}}{\sqrt{|\mathcal{B}_i|}}\\
+\frac{1}{2}\sum_{i=1}^\ec\sw_{i}d(\dd_i,\dd)+\sum_{i=1}^\ec\sw_{i}\nu_i 
\end{aligned} \label{the1-eq}
\end{align}
where $\nu_i=\inf_{h} \gl_{\mathcal{D}_i}(h)+\gl_{\mathcal{D}}(h)$ and $d(\mathcal{D}_i,\mathcal{D})$ measures the distribution discrepancy between two distributions.
\label{the1}
\end{theorem}
The proof is deferred to \Cref{app:genbound}. In \Cref{the1}, the first, second, and third terms in the upper bound show that the generalization performance of the ensemble transfer worsens by the following qualities of each clients: i) bad local model quality on its own training data, ii) small training dataset size, and iii) large discrepancy between its data distribution $\mathcal{D}_i,i\in[K]$ and server's target data distribution $\mathcal{D}$. \fedet~aims in giving lower weights to the clients that demonstrate i) and iii) by weighted consensus distillation where the confidence levels and multiple inferences of the clients contribute to the consensus, so that erroneous outliers can be filtered out. The effect of ii) is also considered in \fedet~by sampling clients in proportion to their dataset sizes. Next, we show through experiments that \fedet~indeed improves the generalization performance of the server model.

\begin{table*}[!t] \centering 
\caption{Best test accuracy achieved by \fedet~and baselines with varying data-heterogeneity. The large server model is used for evaluation for the model homogeneous baselines and the model heterogeneous baselines that require separate server models.}
\vspace{-0.5em}
\begin{tabular}{@{}llccccc@{}}\toprule \label{fig:table2}
& & \multicolumn{2}{c}{$\alpha=0.1$ (Higher Data-Het.)} & \multicolumn{2}{c}{$\alpha=0.5$ (Lower Data-Het.)} & N/A \\
 \cmidrule(lr){3-4} \cmidrule(lr){5-6} \cmidrule(lr){7-7}
& Method &  {CIFAR10} & {CIFAR100} & {CIFAR10} & {CIFAR100} & Sent140 \\ \Xhline{1.5\arrayrulewidth}
\multirow{4}{*}{\begin{tabular}{c}Model\\Homogeneous\\
                \end{tabular}} & FedAvg &  $71.19~\textcolor{darkgray}{(\pm0.27)}$ &   $30.21~\textcolor{darkgray}{(\pm0.32)}$ & $74.82~\textcolor{darkgray}{(\pm0.23)}$ & $33.12~\textcolor{darkgray}{(\pm0.13)}$ & $71.51~\textcolor{darkgray}{(\pm0.45)}$ \vspace*{-0.1em} \\ \cline{2-7} 
& FedProx &  $72.45~\textcolor{darkgray}{(\pm0.13)}$ &  $31.51~\textcolor{darkgray}{(\pm0.11)}$ & $75.24~\textcolor{darkgray}{(\pm0.19)}$ & $33.63~\textcolor{darkgray}{(\pm0.08)}$ & $71.32~\textcolor{darkgray}{(\pm0.31)}$  \vspace*{-0.1em} \\  \cline{2-7} 
& Scaffold &  $75.12~\textcolor{darkgray}{(\pm0.20)}$ &  $30.61~\textcolor{darkgray}{(\pm0.57)}$ & $78.69~\textcolor{darkgray}{(\pm0.15)}$  & $34.91~\textcolor{darkgray}{(\pm0.61)}$ & $73.28~\textcolor{darkgray}{(\pm0.35)}$  \vspace*{-0.1em} \\  \cline{2-7} 
& MOON &  $75.68~\textcolor{darkgray}{(\pm0.51)}$ &  $33.72~\textcolor{darkgray}{(\pm0.89)}$ & $\textbf{81.17}~\textcolor{darkgray}{(\pm0.41)}$ & $\textbf{42.15}~\textcolor{darkgray}{(\pm0.72)}$ & N/A \vspace*{-0.1em} \\  \Xhline{1.75\arrayrulewidth}
\multirow{3}{*}{\begin{tabular}{c}Model\\Heterogeneous\\\end{tabular}}  & FedDF &  $73.81~\textcolor{darkgray}{(\pm0.42)}$ &  $31.87~\textcolor{darkgray}{(\pm0.46)}$ & $76.55~\textcolor{darkgray}{(\pm0.32)}$ & $37.87~\textcolor{darkgray}{(\pm0.31)}$ & $72.19~\textcolor{darkgray}{(\pm0.43)}$  \vspace*{-0.1em} \\  \cline{2-7}  
&DS-FL &  $65.27~\textcolor{darkgray}{(\pm0.53)}$ & $29.12~\textcolor{darkgray}{(\pm0.51)}$ & $68.44~\textcolor{darkgray}{(\pm0.47)}$ & $33.56~\textcolor{darkgray}{(\pm0.55)}$ & $63.12~\textcolor{darkgray}{(\pm0.71)}$  \vspace*{-0.1em} \\  \cline{2-7} 
&Fed-ET (ours) &  $\textbf{78.66}~\textcolor{darkgray}{(\pm0.31)}$  & $\textbf{35.78}~\textcolor{darkgray}{(\pm0.45)}$ & $81.13~\textcolor{darkgray}{(\pm0.28)}$ & $41.58~\textcolor{darkgray}{(\pm0.36)}$ & $\textbf{75.78}~\textcolor{darkgray}{(\pm0.39)}$  \vspace*{-0.1em}\\
\bottomrule  \label{tab:testacc} 
\end{tabular}
\vspace*{-1.5em}
\end{table*}

\section{Experiments}
\label{sec:experiments}
For all experiments, partial client participation is considered where 10 clients are sampled from the 100 clients for image tasks and the 106 clients for the language task. Additional details and results are deferred to \Cref{app:exp}.
\paragraph{Datasets.} For image datasets, the training dataset is partitioned data heterogeneously amongst a total of 100 clients using the Dirichlet distribution $\text{Dir}_{K}(\alpha)$ \cite{hsu2019noniid}. The public dataset is generated by applying a different data transformation to the data samples (non-overlapping with either the training or test dataset) to further differentiate it with the training dataset. For the language task, we use sentiment classification with Sent140 (Twitter) dataset. For the training dataset, users with more than 100 data samples are treated as the FL clients, leading to a total of 106 clients. For all datasets, non-overlapping users' data samples are used.

\paragraph{Models.} For image tasks, we set a CNN, ResNet8, and ResNet18~\cite{he2016deep} for the small server models, and a VGG19~\cite{simonyan2014very} for the large server model. For language tasks, a Tiny-BERT~\cite{bhargava2021tinybert} and a LSTM classifier are set for the small server models, and a Mini-BERT~\cite{bhargava2021tinybert} is set for the large server model. For the representation layers we use a small MLP with dimension 128 which is a small increase in the model size. The small server models in $\mathcal{M}$ are designated (prior to training) to the clients uniformly at random. 

\begin{table}[!t] \centering 
\caption{Communication cost to achieve the target test accuracy $x$ (i.e., $C_{\text{acc}}(x)$) for \fedet~and model homogeneous baselines with the large server model for $\alpha=0.1$.}
\vspace{-0.5em}
\begin{tabular}{@{}lccc@{}}\toprule 
&  {CIFAR10} & {CIFAR100} & Sent140 \\ \midrule
Method  & $C_{\text{acc}}(70\%)$ & $C_{\text{acc}}(30\%)$ & $C_{\text{acc}}(70\%)$ \\ \midrule
FedAvg & $72\times10^9$  & $87\times10^9$ & $25\times10^9$ \vspace*{-0.1em} \\ \hline 
FedProx & $70\times10^9$ & $86\times10^9$ & $22\times10^9$ \vspace*{-0.1em} \\  \hline 
Scaffold & $68\times10^9$ & $79\times10^9$ & $19\times10^9$ \vspace*{-0.1em} \\  \hline 
MOON & $75\times10^9$ & $90\times10^9$ &  N/A \vspace*{-0.1em} \\  \hline
Fed-ET (ours) & $\mathbf{26\times10^9}$ & $ \mathbf{31\times10^9}$ & $\mathbf{9\times10^9}$ \\
\bottomrule  \label{tab:testcomp}
\end{tabular}
\vspace*{-1.7em}
\end{table}

\paragraph{Baselines.} We consider two types of baselines: i) model homogeneous (FedAvg, FedProx, Scaffold, MOON) and ii) model heterogeneous (FedDF, DS-FL). FedGKT assumes full client participation, thus a direct comparison with \fedet~is not possible. Nevertheless, we adapted it to our setup with results presented in~\Cref{app:FedGKT}. For model homogeneous, we use the large server model for evaluation. For model heterogeneous we use the small server models for the client models and the large server model for the server model (if a separate server model is required). We run all experiments with 3 different random seeds, with the variance in the parentheses.

\paragraph{Effectiveness of \fedet.} In \Cref{tab:testacc}, we show the best achieved test accuracy of \fedet~and the baselines for different degrees of data-heterogeneity. \fedet~achieves higher test accuracy for CIFAR10 with high data-heterogeneity ($\alpha=0.1$) and Sent140 compared to both the model homogeneous and model heterogeneous baselines. Specifically, for $\alpha=0.1$, MOON achieves $75\%$ and $33\%$ for CIFAR10 and CIFAR100 respectively at the cost of communicating directly the large VGG19 while \fedet~achieves higher accuracy of $78\%$ and $35\%$ respectively while using smaller models than VGG19 for the clients. For lower data-heterogeneity ($\alpha=0.5$), MOON slightly out-performs \fedet~by around $1\%$ but at the cost of training larger models at the clients. 

\paragraph{Communication Efficiency.} The communication efficiency of \fedet~is shown in \Cref{tab:testcomp}. We compare the communication cost $C_{\text{acc}}(x)$, the total number of model parameters communicated between the server and clients (uplink and downlink) during training to achieve test accuracy $x$. The baselines in \Cref{tab:testcomp} require model-homogeneity, and hence communicate the large server model, while \fedet~communicates the smaller models in $\mathcal{M}$ for each round. \fedet~is able to achieve the target test accuracy with approximately $3\times$ less number of communicated parameters compared to those of the baselines. \fedet~enables efficient training with smaller models at the clients, while achieving comparable performance to when large models are used at the clients.

\paragraph{Effect of the Diversity Parameter $\lambda$.} In \Cref{tab:divcomp}, we show the performance of \fedet~with different values of $\lambda$, which modulates the diversity regularization term in \cref{eq-1-2}. With $\lambda=0$, \fedet~only uses the weighted consensus to train the large server model without leveraging the diversity across the clients' models. A larger $\lambda$ indicates larger regularization loss to include more diversity across the clients' models. For image tasks the best performance is achieved with $\lambda=0.05$, indicating that diversity indeed helps in improving generalization of the server model when moderately applied to the training. For the language task, a larger $\lambda=0.5$ achieves the best performance, demonstrating that depending on the task, more inclusion of the diversity across the models in the ensemble can increase the generalization performance.

\begin{table}[!t] \centering 
\caption{Effect of diversity regularization in \fedet~on test accuracy with different values of $\lambda$ for $\alpha=0.1$.}
\addtolength{\tabcolsep}{-2.5pt}
%
\vspace{-0.5em}
\begin{tabular}{@{}lcccc@{}}\toprule 
&  \multicolumn{3}{c}{Diversity Parameter ($\lambda$)} \\
\cmidrule(lr){2-4} 
Datasets  & $0$ & $0.05$ & $0.5$    \\ \Xhline{1.5\arrayrulewidth}
CIFAR10 &  $76.55~\textcolor{darkgray}{(\pm0.25)}$          
        &$\textbf{78.66}~\textcolor{darkgray}{(\pm0.31)}$ \vspace*{-0.1em}& $75.29~\textcolor{darkgray}{(\pm0.31)}$ \\ \hline 

CIFAR100 &  $31.71~\textcolor{darkgray}{(\pm0.43)}$   &$\textbf{35.78}~\textcolor{darkgray}{(\pm0.45)}$ \vspace*{-0.1em}&$30.18~\textcolor{darkgray}{(\pm0.55)}$ \\  \Xhline{1.75\arrayrulewidth} 
Sent140 & $72.11~\textcolor{darkgray}{(\pm0.28)}$ & $74.37~\textcolor{darkgray}{(\pm0.42)}$  & $\textbf{75.78}~\textcolor{darkgray}{(\pm0.39)}$\vspace*{-0.1em} \\ 
\bottomrule  \label{tab:divcomp}
\end{tabular}
\vspace*{-1.7em}
\end{table}

\section{Conclusion}
Motivated by the rigid constraint of deploying identical model architectures across the clients/server in many FL algorithms, we propose \fedet, an ensemble knowledge transfer framework to train large server models with smaller models trained on clients. Without additional overhead at the clients, \fedet~transfers knowledge to the target model with a data-aware weighted consensus distillation from an ensemble of models trained on heterogeneous data. \fedet~achieves high test accuracy with significantly lower communication overhead and robustness against data-heterogeneity. Relevant future steps are evaluating different deploying strategies of heterogeneous models to the clients and extending \fedet~to a general ensemble knowledge transfer framework. 

\newpage
\bibliographystyle{named}
\bibliography{dist_sgd}
\newpage
\appendix
\section{Proof for \Cref{the1}} \label{app:genbound}
Before presenting the proof, we present several useful lemmas. 

\begin{lemma}[\textbf{Domain adaptation} \cite{dav2009domain}]
With two true distributions $\mathcal{D}_A$ and $\mathcal{D}_B$, for $\forall~\delta\in(0,1)$ and hypothesis $\forall h\in\mathcal{H}$, with probability at least $1-\delta$ over the choice of samples, there exists:
\begin{align}
    \gl_{\mathcal{D}_A}(h)\leq\gl_{\mathcal{D}_B}(h)+\frac{1}{2}d(\mathcal{D}_A,\mathcal{D}_B)+\nu
\end{align}
where $d(\mathcal{D}_A,\mathcal{D}_B)$ measures the distribution discrepancy between two distributions~\cite{dav2009domain} and $\nu=\inf_{h} \gl_{\mathcal{D}_A}(h)+\gl_{\mathcal{D}_B}(h)$.\label{lem3}
\end{lemma}

\begin{lemma}[\textbf{Generalization with limited training samples}]
For $\forall~\ic\in[\ec]$, with probability at least $1-\delta$ over the choice of samples, there exists:
\begin{align}
    \gl_{\dd_\ic}(h_{\ddh_\ic})\leq \gl_{\ddh_\ic}(h_{\ddh_\ic})+\sqrt{\frac{\log{2/\delta}}{2\ldat}}
\end{align}
where $\ldat$ is the number of training samples of client $\ic$. This lemma shows that for small number of training samples, i.e., small $\ldat$, the generalization error increases due to the discrepancy between $\dd_\ic$ and $\ddh_\ic$.
\begin{proof}
We seek to bound the gap between \(\mathcal{L}_{\mathcal{D}_k}(h_{\hat{\mathcal{D}}_k})\) and \(\mathcal{L}_{\hat{\mathcal{D}}_k}(h_{\hat{\mathcal{D}}_k})\). Observe that \(\mathcal{L}_{\mathcal{D}_k}(h_{\hat{\mathcal{D}}_k}) = \mathbb{E}\left[\mathcal{L}_{\hat{\mathcal{D}}_k}(h_{\hat{\mathcal{D}}_k})\right]\), where the expectation is taken over the randomness in the sample draw that generates \(\hat{\mathcal{D}}_k\), and that \(\mathcal{L}_{\hat{\mathcal{D}}_k}(h_{\hat{\mathcal{D}}_k})\) is an empirical mean over losses \(l(h(x), y)\) that lie within \([0,1]\). Since we are simply bounding the difference between a sample average of bounded random variables and its expected value, we can directly apply Hoeffding's inequality to obtain
\begin{align}
    \mathbb{P}\left[\mathcal{L}_{\hat{\mathcal{D}}_k}(h_{\hat{\mathcal{D}}_k}) - \mathcal{L}_{\mathcal{D}_k}(h_{\hat{\mathcal{D}}_k})  \geq \epsilon \right] \leq 2e^{-2m\epsilon^2}.
\end{align}
Setting the right hand side to \(\delta\) and rearranging gives the desired bound with probability at least \(1 - \delta\) over the choice of samples:
\begin{align*}
    \mathcal{L}_{\mathcal{D}_k}(h_{\hat{\mathcal{D}}_k}) & \leq \mathcal{L}_{\hat{\mathcal{D}_k}}(h_{\hat{\mathcal{D}}_k}) + \sqrt{\frac{\log 2/\delta}{2m_k}}.
\end{align*}
\end{proof} \label{lem4}
\end{lemma}
We now present the generalization bound for $\gl_{\dd}\left(\sum_{i=1}^\ec\sw_{i}h_{\ddh_i}\right)$ as follows:
\begin{align}
\begin{aligned}
    &\gl_{\dd}\left(\sum_{i=1}^\ec\sw_{i}h_{\ddh_i}\right)\leqt_{(c)}\sum_{i=1}^\ec\sw_{i}\gl_{\dd}(h_{\ddh_i})\\
    &\leqt_{(d)}\sum_{i=1}^\ec\sw_{i}[\gl_{\dd_i}(h_{\ddh_i})+\frac{1}{2}d(\dd_i,\dd)+\nu_i] \label{eq1-0}
    \end{aligned}
\end{align}
where $\nu_i=\inf_{h} \gl_{\mathcal{D}_i}(h)+\gl_{\mathcal{D}}(h)$, (c) is due to the convexity of $\gl$, and (d) is due to \Cref{lem3}. We can further bound \cref{eq1-0} using \Cref{lem4} as
\begin{align}
&\begin{aligned}
&\gl_{\dd}\left(\sum_{i=1}^\ec\sw_{i}h_{\ddh_i}\right)\leq\sum_{i=1}^\ec\sw_{i}\gl_{\ddh_i}(h_{\ddh_i})\\
&+\sum_{i=1}^\ec\sw_{i}\sqrt{\frac{\log{2/\delta}}{2|\mathcal{B}_i|}}+\frac{1}{2}\sum_{i=1}^\ec\sw_{i}d(\dd_i,\dd)+\sum_{i=1}^\ec\sw_{i}\nu_i
\end{aligned}
    \\
&\begin{aligned}
&=\sum_{i=1}^\ec\sw_{i}\gl_{\ddh_i}(h_{\ddh_i})+\sqrt{\log{\delta^{-1}}}\sum_{i=1}^\ec\frac{\sw_{i}}{\sqrt{|\mathcal{B}_i|}}\\
&+\frac{1}{2}\sum_{i=1}^\ec\sw_{i}d(\dd_i,\dd)+\sum_{i=1}^\ec\sw_{i}\nu_i \label{eq1-1}
\end{aligned}
\end{align}
With \cref{eq1-1}, we finish our proof for \Cref{the1}.

\begin{figure*}[!t]
\centering
\minipage{0.5\textwidth}
\includegraphics[width=3.45in]{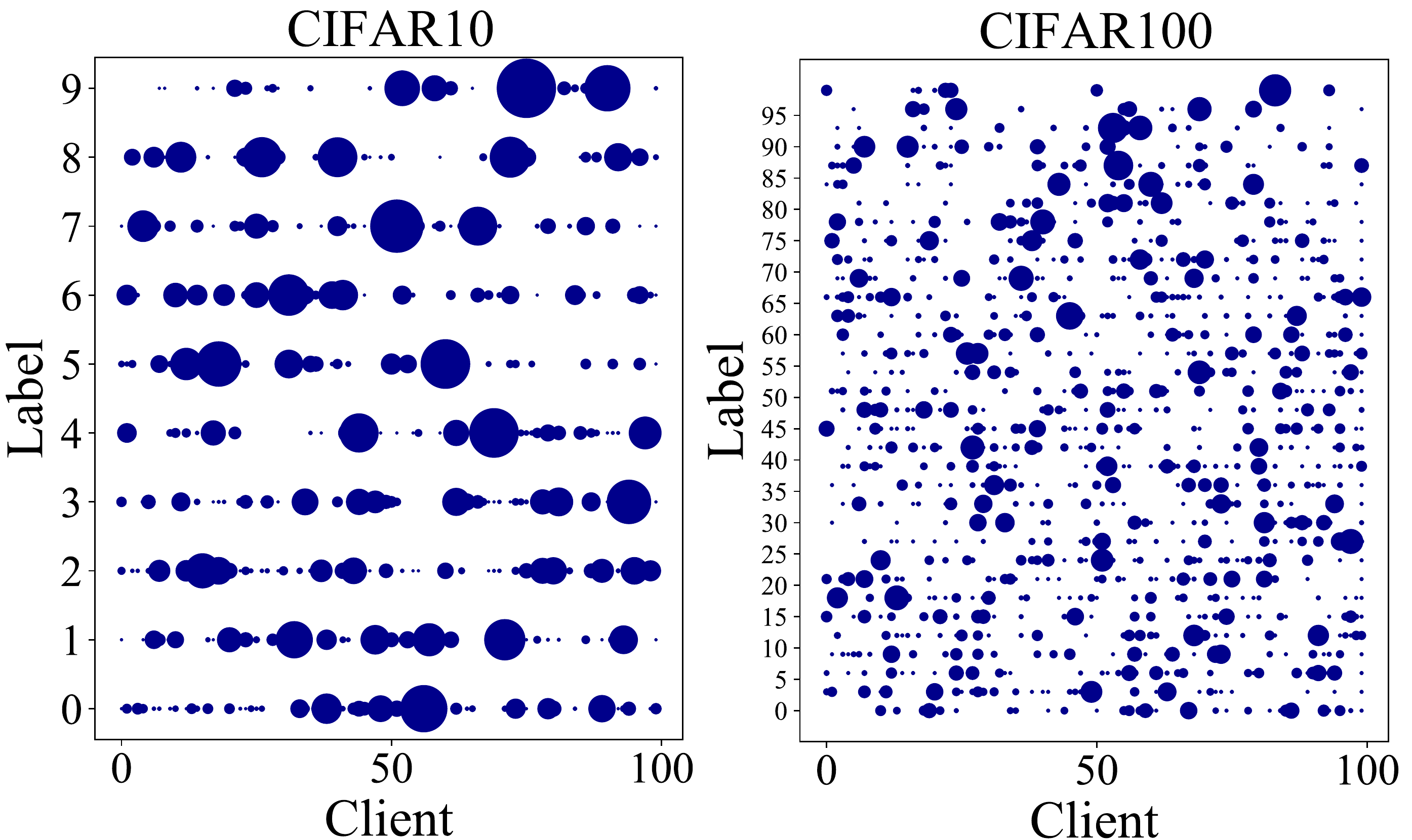}
\caption*{(a): $\alpha=0.1$}
\endminipage\hfill
\minipage{0.5\textwidth}
\includegraphics[width=3.45in]{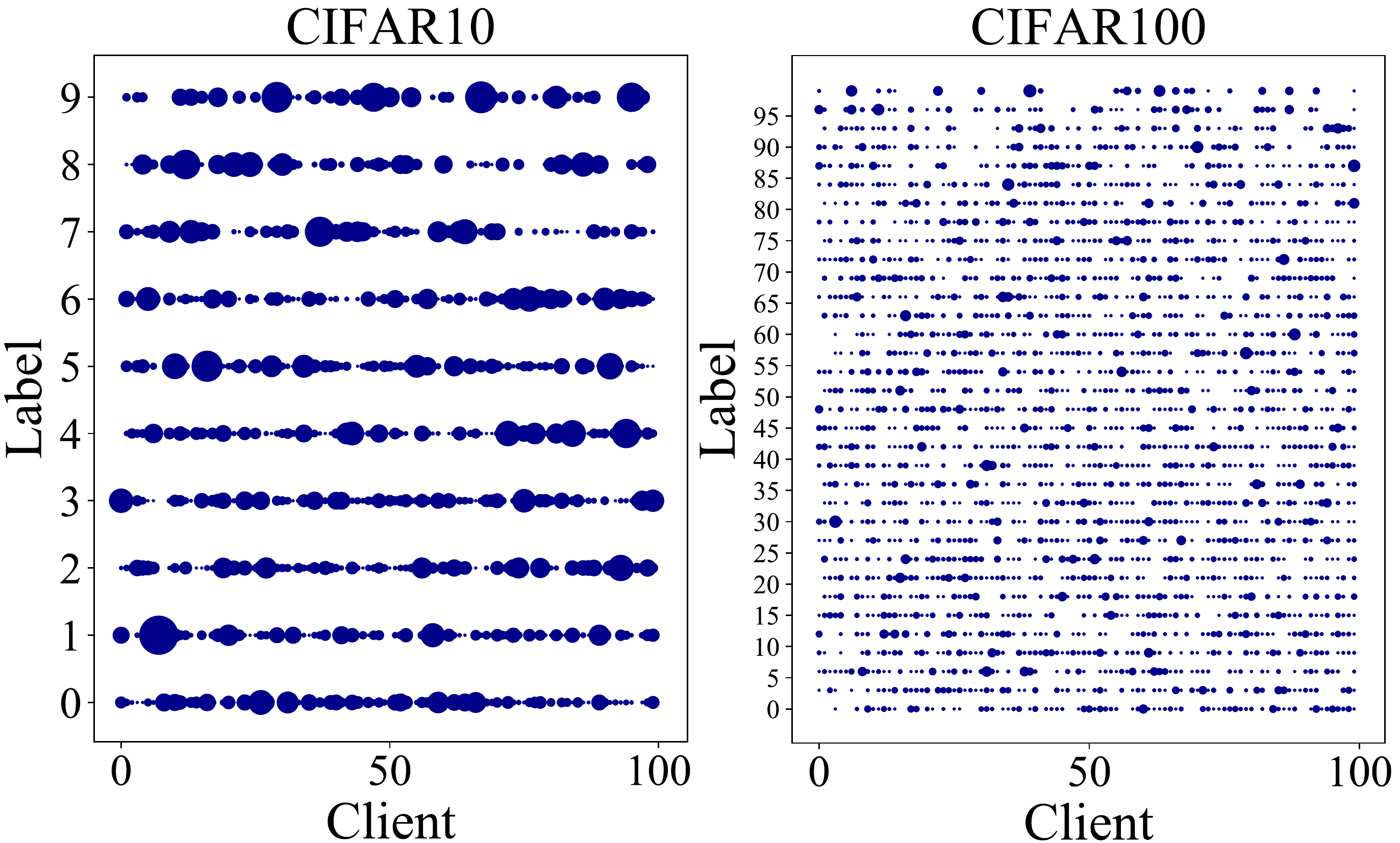}
\caption*{(b): $\alpha=0.5$}
\endminipage\hfill
\caption{Data-distribution with varying $\alpha\in\{0.1,~0.5\}$ for all clients where larger circle indicates larger dataset size for each label 0-9 of CIFAR10 and label 0-99 of CIFAR100.}
 \label{fig:clidetail} 
\end{figure*}

\section{Comparisons with FedGKT} \label{app:FedGKT}

An approach similar to \fedet~for federating heterogeneous models is FedGKT~\cite{chao2020GKT}. FedGKT, however, is presented and implemented in a setting different from ours: the amount of clients is much smaller (experiments presented in the paper are for 16 clients) and, importantly, each client participates in every round. In contrast, we have a large pool of clients from which, at every round, we sample a few. 

Adapting FedGKT to our setting requires significant changes, in particular: i) server logits required for computing the client's loss can no longer be reused and have to be computed on the fly, and ii) client models have to be offloaded to disk or memory, and loaded again when the client is selected. \textbf{This adaptation is very different in spirit from the original FedGKT, as client models are now updated at random intervals.}

Nevertheless, we have implemented this adapted version, starting from the original code, and ran experiments for the setting presented in the main text: 10 clients are picked at random from a pool of 100, which have been assigned partitions of CIFAR-10, sampled using $\alpha \in \{0.1, 0.5\}$. Results are presented in \Cref{tab:fedGKT}. From \Cref{tab:fedGKT} we can see that FedGKT fails to be robust against high-data heterogeneity and partial-client participation compared to \fedet~where the performance gap is particularly high for $\alpha=0.1$.

\begin{table}[!h] \centering 
\caption{Test accuracy for \fedet~and FedGKT on CIFAR-10, with partitions created using both $\alpha=0.1$ and $\alpha=0.5$.}
\begin{tabular}{lcc}\toprule
   Method    & $\alpha = 0.1$      & $\alpha = 0.5$ \\ \midrule
FedGKT & $47.4 (\pm 2.2)$  & $73.2 (\pm 1.2)$ \vspace*{-0.1em}  \vspace*{-0.1em} \\ \midrule
\fedet~(ours) & $\textbf{78.66} (\pm0.31)$ & $\textbf{81.13} (\pm0.28)$ \\ \bottomrule \vspace{-1em} \label{tab:fedGKT}
\end{tabular}
\end{table}

Hyperparameters for FedGKT are similar to those used in the original paper, in the non-i.i.d. case; number of epochs in the server starts at 20 and decreases with time. Multiple variations have been attempted, of batch size, learning rate, learning rate scheduler, and amount of epochs on the server; we report the best results obtained. Moreover, results reported are for the best accuracy during each run, and not necessarily the final one. As in previous experiments, we show an average of 3 runs with different partitions.

\section{Details of Experiment Setup} \label{app:exp}
\paragraph{Dataset.} For the image tasks, the dataset was split to the training/public/test dataset by proportion of 7:1:2, where we split the training dataset across 100 clients by the Dirichlet distribution $\text{Dir}_{K}(\alpha)$ \cite{hsu2019noniid} with $\alpha=0.1$ and $\alpha=0.5$. A smaller $\alpha$ leads to higher data size imbalance and degree of label skew across clients. This is further visualized in \Cref{fig:clidetail} where for larger $\alpha$ we have a more uniform distribution of dataset sizes and labels acros clients. We experiment with three different seeds for the randomness in the dataset partition across clients and present the averaged results across the seeds with the standard deviation. For the public dataset, the following data-augumentation is applied to the allocated dataset so that the public dataset is more differentiated from the training dataset:
\begin{verbatim}
# Transformation for Public-Dataset
from torchvision import transforms as trs
trs.RandomCrop(img_size, padding=4),
trs.RandomHorizontalFlip(),
trs.ColorJitter(0.8*s, 0.8*s, 0.8*s,
    0.2*s) #s=1.0
trs.RandomApply([color_jitter], p=0.8)
trs.RandomGrayscale(p=0.2)
trs.ToTensor(),
trs.Normalize(mean=[0.5, 0.5, 0.5], 
    std=[0.5, 0.5, 0.5])
\end{verbatim}

For the language tasks, the Sent140 dataset was preprocessed to remove users with less than $100$ sentences for the training dataset. The remaining users have been split in $3$ different non-overlapping sets for training, testing and public dataset. The latter dataset is carved out of the training set by removing $20\%$ of the users. Max-sequence length is set to 256 for all data samples. 

\paragraph{Model Setup.} For the image model configuration, for the CNN we have a self-defined convolutional neural network with 2 convolutional layers with max pooling and 4 hidden fully connected linear layers of units $[120,100,84,50]$ with the representation layer in the end. The input is the flattened convolution output and the output is consisted of 10 or 100 units each of one of the 0-9 labels or 0-99 labels. For the VGG and the ResNets, we use a modified version of the open-sourced VGG net and ResNets from Pytorch (torchvision ver.0.4.1) with pretrained, batchnorm as both False with the representation layer in the end. For the language model configuration, we use the pretrained Tiny-BERT and pretrained Mini-BERT open sourced through huggingface~\cite{bhargava2021tinybert} and a self-defined two layer LSTM binary classifier with 256 hidden units connected with the representation layer in the end. The number of parameters for the used models in this work is shown in \Cref{tab:modelsize}.

\begin{table}[!h] \centering 
\caption{Number of Trainable Parameters for Different Models used in the experimental setup with Unit as M=$10^6$}
\begin{tabular}{@{}lccc@{}}\toprule 
Task & Server Model &  Models in $\mathcal{M}$ \\ \midrule
\multirow{3}{*}{Image} &  & Res18 ($11.5$M) \\ 
                    &  VGG19 ($20.9$M) & Res8 ($5.2$M)  \\ 
                    &  & CNN ($0.4$M) \\ 
\Xhline{1.75\arrayrulewidth}
\multirow{2}{*}{Language} & \multirow{2}{*}{Mini-BERT~($11.3$M)} & Tiny-BERT ($4.5$M) \\ & & LSTM ($2.5$M)  \\ 
\bottomrule \label{tab:modelsize}
\end{tabular}
\end{table}

\paragraph{Hyperparameters and Training.} All algorithms are ran until convergence on the validation dataset. For the image tasks' local-training hyperparameters, we do a grid search over the learning rate: $\eta\in\{0.1, 0.05, 0.01, 0.005, 0.001\}$, batch-size: $b\in\{32, 64, 128\}$, and local iterations: $\tau\in\{10, 30, 50\}$ to find the hyper-parameters with the highest test accuracy for each benchmark. For fair comparison, we do not use learning rate decay. For all benchmarks we use the best hyper-parameter for each benchmark after doing a grid search over feasible parameters referring to their source codes that are open-sourced. For the server-side hyperparameters for image tasks, we do a grid search over the learning rate: $\eta_s\in\{0.1, 0.05, 0.01, 0.005, 0.001\}$, the public batch size: $b_s\in\{32, 64, 128, 256\}$, SGD iterations: $\tau_s\in\{10, 30, 50\}$, and regularization weight $\lambda\in\{0, 0.05, 0.5\}$ to find the best working hyperparameters. The best hyperparameters used for image tasks are $\eta=0.1,b=64,\tau=30,\eta_s=0.005,b_s=64,\tau_s=128,\lambda=0.05$. For the language tasks' local-training hyperparameters we use the best hyperparameters for the Sent140 with tiny-BERT as $\eta=0.0003,b=5,\tau=15$. For the server-side hyperparameters, we do a grid search over the learning rate: $\eta_s\in\{0.1, 0.01, 0.001\}$, the public batch size: $b_s\in\{5, 10, 15\}$, SGD iterations: $\tau_s\in\{10, 15, 20\}$, and regularization weight $\lambda\in\{0, 0.05, 0.5\}$ to find the best working hyperparameters. The best hyperparameters used are $\eta_s=0.001,b_s=5,\tau_s=10,\lambda=0.5$.

\paragraph{Platform.} All experiments are conducted with clusters equipped with one NVIDIA TitanX GPU. The number of clusters we use is fixed to the fraction of clients we select. The machines communicate amongst each other through Ethernet to transfer the model parameters and information necessary for client selection. Each machine is regarded as one client in the federated learning setting. The algorithms are implemented by PyTorch.


\end{document}